%% file: main.tex
\documentclass{article}

\usepackage{PRIMEarxiv}

\usepackage[utf8]{inputenc} 
\usepackage[T1]{fontenc}    
\usepackage[hidelinks]{hyperref}       
\usepackage{url}            
\usepackage{booktabs}       
\usepackage{array}          
\usepackage{amsfonts}       
\usepackage{nicefrac}       
\usepackage{microtype}      
\usepackage{lipsum}
\usepackage{amsmath}
\usepackage{fancyhdr}       
\usepackage{graphicx}       
\usepackage{xcolor}         
\usepackage{todonotes}
\usepackage[round]{natbib}
\usepackage[title]{appendix}
\graphicspath{{media/}}     

\title{Linear representations of grammaticality in neural language models}

\author{
  Jane Li\\
  Department of Cognitive Science \\
  Johns Hopkins University \\
  Baltimore, MD\\
  \texttt{sli213@jhu.edu} \\
   \And
  Najoung Kim \\
  Department of Linguistics \\
  Boston University \\
  Boston, MA\\
  \texttt{najoung@bu.edu} \\
}

\newcounter{daggerfootnote}
\newcommand*{\daggerfootnote}[1]{%
    \setcounter{daggerfootnote}{\value{footnote}}%
    \renewcommand*{\thefootnote}{\fnsymbol{footnote}}%
    \footnote[2]{#1}%
    \setcounter{footnote}{\value{daggerfootnote}}%
    \renewcommand*{\thefootnote}{\arabic{footnote}}%
    }

\begin{document}
\maketitle

\begin{abstract}
Whether neural language models (NLMs) possess the ability to distinguish strings on the basis of their grammaticality remains a debated topic in the computational linguistics literature. Existing evidence has largely relied on probability-based measures, testing whether models assign higher probabilities to grammatical than ungrammatical strings. However, probability comparisons have been criticized as a measure for grammatical knowledge based on the assumption that grammaticality is inherently entangled with likelihood. Model-assigned probability is a function of many related sentence properties, such as lexical frequency, plausibility, and world knowledge. In this work, we move beyond probability-based evaluations and directly investigate whether grammaticality is encoded in the internal representations of NLMs. Using linear probing (specifically, mass-mean probing), we test whether grammatical and ungrammatical sentences are systematically separated in representational space: i.e., whether this distinction can be detected even by a low-complexity classifier. We further examine the extent to which these representations are independent of sentence properties that are correlated with grammaticality, as well as their generalization across grammatical phenomena and languages. Our results provide evidence that grammaticality is robustly encoded in sentence representations of a wide range of pretrained NLMs, yielding clear representational separation between grammatical and ungrammatical sentences that cannot be fully explained by alternative sentence-level factors. Moreover, this encoding generalizes across a broad range of grammatical phenomena and to some degree, across languages, suggesting that grammaticality constitutes a coherent representational dimension in contemporary NLMs. These findings contribute new evidence to debates about the nature of syntactic knowledge in language models and offer a complementary framework for evaluating grammatical competence that is not dependent on string probabilities alone.\daggerfootnote{Code, analyses, and data are available at \hyperlink{https://github.com/jane-lisy/linear-gram}{https://github.com/jane-lisy/linear-gram}.}
\end{abstract}

\section{Introduction}
The rise of neural language models (NLMs), trained to predict the likelihood of vast amounts of text, has led to a resurgence in a core debate in cognitive science about whether statistical learners can distinguish grammaticality of strings \citep{chomsky1957SyntacticStructures,pereira2000Formalgrammarinformation,katzir2023WhyLargeLanguage,dentella2023Systematictestingthree,warstadt2020BLiMPBenchmarkLinguistic,hu2026WhatCanString}. On the one hand, many pretrained NLMs assign higher probabilities to unseen grammatical strings over their ungrammatical counterparts \citep{warstadt2020BLiMPBenchmarkLinguistic,hu2026WhatCanString}, supporting arguments that they capture non-trivial aspects of grammar \citep{futrell2025HowLinguisticsLearned} and that this method can serve as a linking function between model-assigned probabilities and grammaticality \citep{hu2026WhatCanString}. There is significant pushback on this view, with one prominent line of disagreement being that because model-assigned probabilities fundamentally entangle grammaticality with likelihood, inferences about model knowledge of grammaticality is fraught \citep{katzir2023WhyLargeLanguage,dentella2023Systematictestingthree}. In this work, we argue that stronger evidence for sensitivity to grammaticality comes from the distinction being reflected in the internal organization of sentence representations, unfettered by various sentence properties that correlate with grammaticality.

Humans reliably distinguish well-formed sentences from ill-formed ones, even when surface statistics are uninformative, suggesting sensitivity to abstract structural constraints \citep{chomsky1957SyntacticStructures,pinker1991RulesLanguage}. In symbolic theories of syntax, and symbolic model instantiations of these theories (e.g., via a context-free grammar), grammaticality as a property of a sentence falls out naturally. What about grammaticality in NLMs? On one view, Chomsky's (1957) demonstration of the orthogonality of probability and grammaticality seems to dissuade us from the idea that a language model -- crucially a probabilisitic model of language -- can have true discernment between grammatical and ungrammatical strings. This position has received significant theoretical and empirical pushback (\citeauthor{pereira2000Formalgrammarinformation}, \citeyear{pereira2000Formalgrammarinformation}; and more recently, \citeauthor{hu2026WhatCanString}, \citeyear{hu2026WhatCanString}). Setting aside this in-principle disagreement, an empirical limitation of probability-based measures of grammaticality in modern LMs lies in the amount of information it encodes. Other string properties such as word frequencies, sentence length \citep{tjuatja2025WhatGoesLM,lau2017GrammaticalityAcceptabilityProbability}, plausibility \citep{lepori2026ThisJustFantasy}, and world knowledge \citep{ivanova2025ElementsWorldKnowledge}, are documented to influence string probability. Even when minimal pair comparisons control for some of these factors (e.g., plausibility), it remains largely unclear how, or whether, these factors interact with grammaticality to yield the types of probability distinctions we observe. These limitations further motivate the study of whether grammaticality is representationally encoded in NLMs that are argued to have grammaticality distinctions.

Zooming out of discussions about NLM grammaticality detection, having a more acute understanding of what syntactic knowledge a computational model possesses -- in the form of sentence grammaticality -- is a crucial precondition for work that argues NLMs can provide insight into human cognition. Arguments that neural language models can provide insight into the human language faculty \citep{futrell2019Neurallanguagemodels,mansfield2025LookingforwardLinguistic} rest on the assumption that we are able to reliably determine whether (a) a model is able to discern grammaticality at all, or (b) a model has grammatical knowledge of particular phenomena. For example, work arguing that certain phenomena are in principle learnable from data alone with little domain-specific biases requires positive or negative demonstrations of the learned grammaticality distinctions \citep{wilcox2024UsingComputationalModels}. On the flip side, based on the cited criticisms of probability-based methods (confounds with likelihood, encoding other correlated factors), related work has argued against the utility of NLMs for the study of human cognitive science. \citep{katzir2023WhyLargeLanguage,fox2024LargeLanguageModels}. Under this view, these models neither have genuine knowledge of grammaticality or syntax, and their successes with regard to fluent text generation or interpretation are surface-level in nature \citep{bender2021DangersStochasticParrots}, thus not meeting the preconditions that make them suitable models for inferences about the human mind. The questions at stake here further highlight the importance of additional evidence to adjudicate between these positions.

It is clear that settling the question of whether NLMs have true distinctions of grammaticality is of timely essence. In this work, we approach this question by moving beyond probability-based measures and probing for grammaticality representations from model representations of sentences. In this article, we provide evidence for the argument that pretrained NLMs have grammaticality representations by demonstrating that there is clear representational separation between grammatical and ungrammatical sentences in a model's representational space. We find that this separation is distinct from many collinear properties of sentences, while being widely generalizable beyond a particular grammatical phenomenon. At a high level, this work addresses three questions:

\begin{enumerate}
    \item[\textbf{RQ1}.]  Is there a detectable encoding of grammaticality in sentence representations? Is it measurable even with a low-complexity (i.e., mass-mean) probe? 
    \item[\textbf{RQ2}.] How is this grammaticality representation distinct from collinear properties of sentences?
    \item[\textbf{RQ3}.] How generalizable is this grammaticality encoding across grammatical phenomena or languages? 
\end{enumerate}

In the next subsections, we review existing methods for grammaticality detection for NLMs and further flesh out criticisms against these methods, to inform us of how a study of grammaticality representation should address these confounds. We also review model interpretability literature to motivate the logic used in our experiments that follow.

\subsection{Previous work}
\subsubsection{Grammaticality detection}
The most common method for grammaticality detection is targeted syntactic evaluation using model-assigned sentence probabilities \citep{marvin2018TargetedSyntacticEvaluation,hu2020SystematicAssessmentSyntactic,warstadt2020BLiMPBenchmarkLinguistic}. Probabilistic models are evaluated on their ability to consistently assign higher probability to grammatical strings over their minimally-different ungrammatical counterparts. There exists an active line of work developing the kinds of sentential comparisons that would be insightful for understanding the syntactic capabilities of the models in question, in English \citep{marvin2018TargetedSyntacticEvaluation,futrell2019Neurallanguagemodels,warstadt2019NeuralNetworkAcceptability,warstadt2020BLiMPBenchmarkLinguistic,hu2020SystematicAssessmentSyntactic,huebner2021BabyBERTaLearningMore,vazquezmartinez2023EvaluatingNeuralLanguage}, Japanese \citep{someya2023JBLiMPJapaneseBenchmark}, Mandarin Chinese \citep{song2022SLINGSinoLinguistic,liu2025SystematicAssessmentLanguage}, Russian \citep{taktasheva2024RuBLiMPRussianBenchmark}, and more \citep{mueller2020CrossLinguisticSyntacticEvaluation,jumelet2025MultiBLiMP10Massively}. As reviewed, while this method has served as indices for grammatical knowledge in many recent work in the natural language processing literature \citep{hu2025CircuitsChomskyPrepretraininga,agarwal2025MechanismsvsOutcomes,yang2026UnifiedAssessmentPoverty}, the assumption that models have genuine distinctions of grammaticality is challenged \citep{dentella2023Systematictestingthree,katzir2023WhyLargeLanguage,leivada2024ReplyHuApplying}.

The primary argument is that ungrammatical sentences are unlikely, entangling the two factors in the training data. With the training objective of NLMs being that of maximizing the likelihood of this entangled data, it is unclear whether probability contrasts are indexing likelihood or grammaticality. \cite{katzir2023WhyLargeLanguage} supports this claim with demonstrations that there exists metalinguistic queries where an LLM, likely succeeding on these probability-based benchmarks, will prefer a likely, but ungrammatical, continuation, over a grammatical continuation. Similar observations were made in \cite{wilcox2024UsingComputationalModels} and \cite{levy2024ScienceLanguageEra}. Thus, a demonstration that models have grammaticality distinctions in representational space would also need to demonstrate that the success of the grammaticality probes is not due to string probability differences or extra-grammatical factors in the fitting sentences.

Another line of research uses metalinguistic prompting to measure a model's sensitivity to string grammaticality \citep{dentella2023Systematictestingthree,behzad2024AskLLMsEnglish}. The target sentence is embedded in a query directing asking about the grammaticality of the sentence, and the measured variable is whether the model responds to this query positively (e.g., by responding ``Yes'') or negatively. This form of evaluation is useful for models where probabilities or weights are not publicly accessible, but makes the additional assumption that models have the ability to provide metalinguistic judgments for a sentence: an ability that is distinct from having a grammaticality distinction. However, \cite{hu2023Promptingnotsubstitute} argue that metalinguistic prompt-based evaluations underestimate a model's syntactic ability.

Finally, we review existing work that probes for model-internal representations of grammaticality. To our knowledge, \citet{wang2026ImplicitRepresentationsGrammaticality} is the only study that probes for grammaticality from a model's representations of sentences. Their work applied linear probes (logistic regression classifiers trained on sentence representations) to differentiate between sentences drawn from corpora versus perturbed sentences that render the sentence ungrammatical, via functions such as word insertion, deletion, or shuffling. Their results provide convergent evidence that grammaticality is robustly represented various LLM's representations of sentences. In our work, we take a step further to establish the generalizability of linear encodings of grammaticality across various kinds of violations. Complementary to probing, \citet{duan2025HowSyntaxSpecialization} investigate how syntactic specialization for specific grammatical phenomena emerges in activation space by comparing hidden state activations elicited by grammatical and ungrammatical sentence pairs. Although their goal is to characterize functional specialization rather than decode grammaticality with probes, their activation space comparisons are closely related to our methodology and likewise provide evidence that syntactic information is systematically reflected in model representations. Aside from probing in hidden state space, there also exists a line of circuitry work that examines the weights that are involved in computations of particular phenomena (e.g., subject-verb agreement) and whether they generalize across languages \citep{zhang2024SameDifferentStructural,kryvosheieva2025Differenttypessyntactic}. These studies demonstrate that a small number of units in a model's weight space are responsible for distinctions in probability space, and its generality provides convergent evidence that LLMs have robust representations of syntactic structure \citep{hewitt2019StructuralProbeFinding,eisape2022ProbingIncrementalParse}.

\subsubsection{Mass-mean probing}
We investigate grammaticality as encodings in sentence representations using mass-mean probing \citep{marks2024GeometryTruthEmergent}, a simple linear probing method with minimal data and modeling assumptions. Mass-mean probing has been used to detect a range of high-level properties in neural representations, including truthfulness \citep{marks2024GeometryTruthEmergent}, sentence plausibility \citep{lepori2026ThisJustFantasy}, and political orientation \citep{kim2025LinearRepresentationsPolitical}, and has also been adopted in model-steering work \citep[Contrastive Activation Addition;][]{rimsky2024SteeringLlama2}. Its appeal lies in its interpretability and low complexity: the probe consists of a single direction in representation space defined by the difference between the mean representations of two classes. In this work, we apply this method to test whether grammaticality corresponds to a stable and accessible dimension of sentence representations in neural language models.

Through our experiments, we test the following hypothesis: if it is the case that a representational encoding of grammaticality exists, we should expect to see cross-phenomena generalization, while expecting to observe no generalization to sentential properties that correlate with grammaticality. We evaluate this hypothesis across 25 publicly available pretrained NLMs that perform above chance on targeted syntactic evaluation benchmarks, thus establishing a prerequisite level of behavioural sensitivity to syntactic structure.

\section{Methods}
\begin{figure}[h!]
    \centering
    \includegraphics[width=\linewidth]{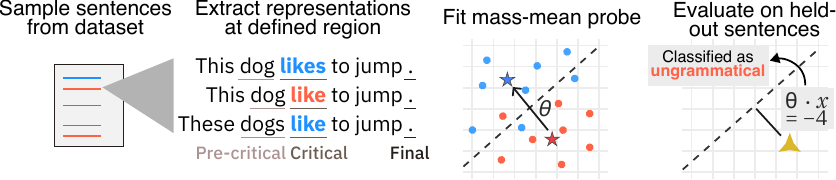}
    \caption{A step-by-step summary of the methods used in the experiments of this paper.}
    \label{fig:method}
\end{figure}
\subsection{Datasets}\label{sec:dataset}
We use datasets and benchmarks from previous work that target grammaticality, as well as other sentential properties for our deconfounding experiments. Tables \ref{tab:gram-dat} and \ref{tab:extragram-dat} summarize these sources respectively. Pre-processing procedures for specific datasets are documented in Appendix \ref{app:dataset}. The grammaticality benchmarks \citep[most notably, BLiMP and SCaMP;][]{warstadt2020BLiMPBenchmarkLinguistic,mccoy2025Modelingrapidlanguage} span a wide range of linguistic phenomena, allowing us to test within- and between-phenomena generalization in our experiments. We note a subset of the BLiMP datasets (40 out of 67) are annotated for the word where grammaticality diverges in a minimal pair setting (e.g., \textit{These dogs} \{\textit{like}, *\textit{likes}\} \textit{to jump.}). We use these annotations (see \S\ref{sec:extract}) to conduct additional analyses on this particular subset of datasets.
\begin{table}[h!]
    \centering
    \begin{tabular}{p{6cm}p{3cm}ll}
    \toprule
        Benchmark & Language & \# Datasets & \# Items per dataset \\
    \midrule
        \textbf{BLiMP} \citep{warstadt2020BLiMPBenchmarkLinguistic} & English & 67 & 2,000 \\
        \textbf{SCaMP} \citep{mccoy2025Modelingrapidlanguage} & English & 134 & 2,000 \\
        \textbf{DGL} \citep{dentella2023Systematictestingthree, hu2024Languagemodelsalign} & English & 1 & 190\\
        \textbf{LI-Adger} \citep{sprouse2012Assessingreliabilitytextbook,sprouse2013comparisoninformalformal,vazquezmartinez2023EvaluatingNeuralLanguage} & English & 1 & 4,066 \\
        \textbf{RUBLiMP} \citep{taktasheva2024RuBLiMPRussianBenchmark} & Russian & 45 & 2,000\\
        \textbf{ZhoBLiMP} \citep{liu2025SystematicAssessmentLanguage} & Mandarin Chinese & 118 & 2,000\\
        \textbf{CLAMS} \citep{mueller2020CrossLinguisticSyntacticEvaluation} & French, German, Hebrew, Russian & 28 & 280--2,000\\
    \bottomrule\\
    \end{tabular}
    \caption{Benchmarks and datasets contrasting sentence grammaticality used in this work.}
    \label{tab:gram-dat}
\end{table}
\begin{table}[h!]
    \centering
    \begin{tabular}{lp{5cm}ll}
    \toprule
        Benchmark & Domain & \# Datasets & \# Items per dataset \\
    \midrule
        \textbf{Shades} \citep{hu2025ShadeszeroDistinguishing} & Event plausibility/conceivability & 6 & 2000 \\
        \textbf{Drive} \citep{tuckute2024Drivingsuppressinghuman} &  Sentence imageability, arousal, etc. & 9 & 1542\\
        \textbf{Truth} \citep{marks2024GeometryTruthEmergent} & Sentence factuality & 11 & 354--2000\\
    \bottomrule\\
    \end{tabular}
    \caption{Benchmarks and datasets contrasting sentence properties that are not grammaticality, but may contribute to sentence acceptability judgments in humans and models (extra-grammatical). All data are in English.}
    \label{tab:extragram-dat}
\end{table}

\subsection{Models}
This work examines the hidden states of 25 publicly available pretrained NLMs. These models vary widely in their scale (14 million to 14 billion parameters) and model family, as listed in Table \ref{tab:model}. All of these models examined perform above chance at discerning grammaticality via minimal pair string probability comparisons on BLiMP (Appendix \ref{app:models}), which we take as the precondition to expect any knowledge of grammaticality to be encoded in each of these models, should it be the case that behavioural success is due to its sensitivity to grammaticality.
\begin{table}[h!]
    \centering
    \begin{tabular}{ll}
        \toprule
        Model family & Models (and parameter count) \\
        \midrule
         GPT2 & Small (117M), Medium (345M), Large (774M), XL (1.5B)\\
         Llama & Llama-3.2-1B, Llama-3.2-3B, Llama-3.1-8B\\
         OLMo & OLMo-1B, OLMo-7B, OLMo-2-1B, OLMo-2-7B, OLMo-2-13B\\
         Pythia & 14M, 70M, 160M, 410M, 1B, 1.4B, 2.8B, 6.9B, 12B\\
         Qwen3 & 0.6B, 1.7B, 4B, 14B\\
         \bottomrule\\
    \end{tabular}
    \caption{The models examined in this work.}
    \label{tab:model}
\end{table}

\subsection{Extracting representations}
\label{sec:extract}
Each sentence is passed into the model verbatim. We perform our analyses over final and intermediate representations of sentences in the final hidden layer. When available, we define three regions-of-interest over tokens.

\begin{itemize}
    \item Critical region: the span of tokens in the sentence where grammaticality divergence becomes clear (i.e., there is no possible grammatical continuation of the ungrammatical sentence). When this span contains more than one token, we take the average across tokens.
    \item Pre-critical region: the token immediately preceding the critical region. This is treated as a control condition, because even though strings may diverge in their content, there are no grammaticality differences between the minimal pairs at this point of the sentence. 
    \item Final region: the final token of the sentence, which is the punctuation mark that ends the sentence (e.g., period or question mark).
\end{itemize}

An example illustrating these regions are available in Figure \ref{fig:method}. When region-level annotations are not made available, the term ``sentence representation'' refers to the representation extracted in the Final region, which is accessible for any sentence. All representations extracted are $d_M$-dimensional vectors, where $d_M$ is the dimension of the final hidden layer of the model $M$. Before conducting any probing analyses, we mean-center all representations. The mean is computed from the set of representations that refer to the same token, and then subtracted from the representation. This is done to ensure that any probing results are not a reflection of the lexical/token identities of tokens in a sentence representation.

\subsection{Mass-mean probing}
We search for grammaticality representations via mass-mean probing \citep{li2023InferenceTimeInterventionEliciting,marks2024GeometryTruthEmergent}, a linear probe with minimal data assumptions. Let there be a set of data points $\mathcal{D} = \{(x_i, y_i)\}$ where $x_i \in \mathbb{R}^d$ (sentence representations) and $y_i \in \{0, 1\}$ (grammaticality labels). The mean representation for each class, $\mu^0$ and $\mu^1$, as well as a difference vector $\theta = \mu^1 - \mu^0$ is computed. A mass-mean probe $g: \mathbb{R}^d \mapsto [0,1]$ is defined as: $g(y = 1 \mid x) = \sigma(\theta^Tx)$. For classification, a representation is classified as 1 (grammatical) when $\sigma(\theta^Tx) > 0.5$, or $\theta^Tx > 0$. The decision boundary for the mass-mean probe is the plane that satisfies the equation $\theta^Tx = 0$.

Unless otherwise specified, a probe is fitted to 1,600 sentences. For each probing setup, we report the results from 100 runs, with each run sampling a different random subset of whatever fitting dataset defined by the experiment. For the remainder of this work, we will use the terms ``fitting dataset'' and ``evaluation dataset'' to refer to the set of sentences where the fitting/training sentences or evaluation/test sentences are drawn from. The next subsection reports how these results are processed and reported.

\subsection{Evaluation, measurements, and significance testing}
\label{methods:eval}
All evaluations are conducted on held-out sentences. Oftentimes, the fitting and evaluation dataset target different phenomena (e.g., fitting on an NPI-licensing dataset and testing on a verb agreement dataset). In that case, we randomly sample fitting subsets across runs, while evaluating on the totality of the dataset for the evaluated phenomenon. In the other scenario, where the experiment involves fitting and evaluating on the same dataset, we treat the held-out remainder of the fitting dataset as the evaluation dataset. For example, for a dataset with 2,000 sentences, the probe is fitted on 1,600 randomly sampled sentences from the dataset, and then evaluated on the held-out 400 sentences.

For each probing setup, we now have a distribution over all probing measures, e.g., item-level accuracy, dataset-level accuracy, an item's distance to classification boundary. Unless otherwise specified, we report the mean of these measures. One crucial second-order measure that we report is \textbf{probe success}, taken to represent whether this probing setup successfully distinguishes grammaticality (or whatever construct that defines the labels) of the evaluated dataset. This is measured by comparing the dataset-level accuracy distribution against an equivalent null distribution. A null probe is constructed in the same way as a normal probe, except the labels of the fitting sentences are shuffled. We consider a probing configuration to be successful if the probing distribution is significantly greater than the null distribution (via a two-sample Welch's $t$-test, $p$-value is Holm-Bonferroni adjusted) and the mean of the probing distribution is at least 5\% greater than the null. This threshold acts as a noise threshold to avoid false positives. Observing that probes fitted on single BLiMP datasets at the Pre-Critical region (a manipulation in Experiment 2) are in the range [-2.62\%, 1.65\%], we set the threshold to 5\% as a conservative estimate of false positives.

\section{Experiment 1: Grammaticality detection via mass-mean probing}
In this initial experiment, we establish that grammaticality distinctions are detectable in hidden state space via mass-mean probing. Demonstrating this basic finding in a given model allows us to then inquire the scope, or sensitivity, of these grammaticality representations in the next experiments. Additionally, to situate this work in the prior literature on grammaticality measurements, we also report our results in a way that yields comparisons between probing and probability-based measures.

We fit a probe on sentences randomly sampled from the LI-Adger dataset \citep{vazquezmartinez2023EvaluatingNeuralLanguage,sprouse2012Assessingreliabilitytextbook,sprouse2013comparisoninformalformal,sprouse2012Assessingreliabilitytextbook}, a collection of acceptability judgment data from linguistic theory papers and a prominent syntax textbook. Then, we evaluate the probe on all BLiMP sentences (67 datasets). We report probe accuracy and probe success, the two main measures used in this paper, but also derive a comparison-based measure, MP Rep (Minimal Pair, Representations), that compares the probabilities assigned by the probe in a similar fashion to targeted syntactic evaluation. For each dataset, we define MP Rep to be the proportion of minimal pairs in the dataset where $g(x_\text{grammatical}) > g(x_\text{ungrammatical})$, $g(x)$ being the mass-mean probe-assigned probability for the positive label. This measure is also used in \cite{wang2026ImplicitRepresentationsGrammaticality} and \cite{duan2025HowSyntaxSpecialization} for comparisons between probe- and probability-based measures.

\subsection{Result \#1: All models exhibit probe sensitivity to grammaticality}
\begin{figure}[h!]
    \centering
    \includegraphics[width=\linewidth]{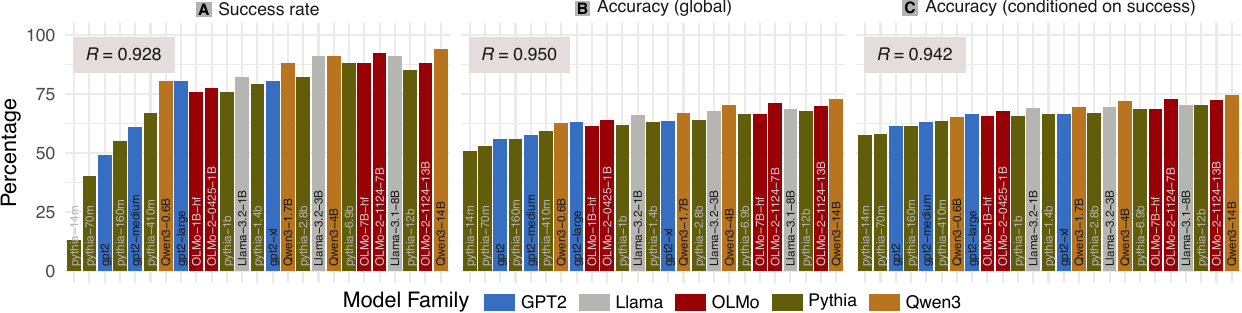}
    \caption{Model-level results from Experiment 1. $R$ reports the Pearson correlation between model size (log10 parameter count) and the main measure of each subplot. (\textbf{A}) Percentage of successful probes out of the 67 BLiMP datasets. (\textbf{B}) Average accuracy (in \%) of each probe. (\textbf{C}) Average accuracy within the set of successful probes.}
    \label{fig:exp0}
\end{figure}
We find that all 25 models we examine demonstrate some probe sensitivity to grammaticality, albeit with great variation between models (Fig.\ \ref{fig:exp0}A). Success rate ranges from 13.4\% in Pythia-14m to 94.0\% in Qwen3-14B. Though, aside from Pythia-14m, all models exhibit probe sensitivity on at least 27 datasets (of 67 datasets total), suggesting that there exists some non-trivial degree of representational separation in each of these models. Model size (in terms of number of parameters) best predicts the variation in success rate (see $R$ values in Fig.~\ref{fig:exp0}) -- we explore the relationship between model size and probing performance in further detail in then next subsections. Global accuracy fluctuates similarly (Fig.\ \ref{fig:exp0}B), though this is expected as accuracy on unresponsive datasets hovers around 50\% for all models, drawing the global accuracy down in proportion to the number of successful datasets for a given model. Narrowing our analyses down to the accuracy of successful probes, we observe a similar effect (Fig.\ \ref{fig:exp0}C).

\subsection{Result \#2: Comparisons between probe and probability-based measures}
\begin{figure}[h!]
    \centering
    \includegraphics[width=\linewidth]{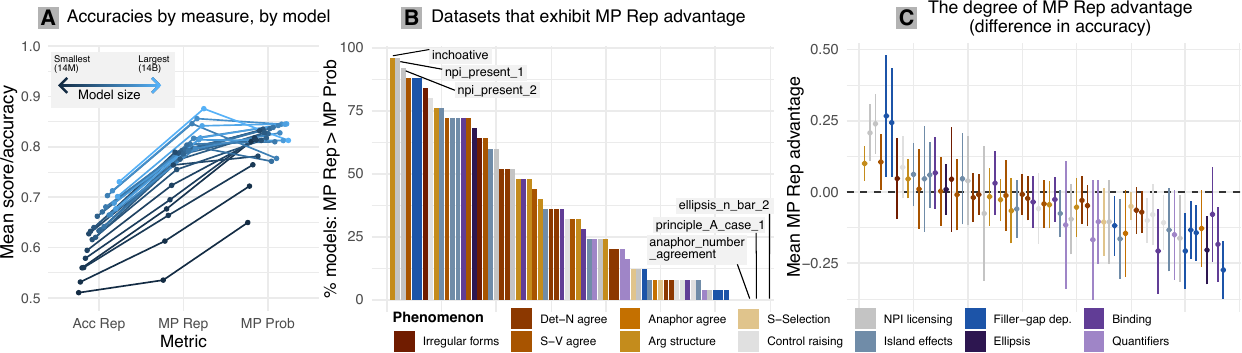}
    \caption{(\textbf{A}) Comparisons between accuracy measures. Each point represents an aggregate measure of accuracy for a model. (\textbf{B}) A dataset-level comparison of MP Rep advantage, ordered on the x-axis by the proportion of models that exhibit advantage on that dataset. (\textbf{C}) The magnitude of the MP Rep advantage, following the ordering as sub-figure \ref{fig:exp0:comp}B.}
    \label{fig:exp0:comp}
\end{figure}
Our results show that probing accuracy calculated from labels determined by a fixed threshold of $g(x)$, in general underperforms relative to minimal-pair, probability-based targeted syntactic evaluation. Fig.\ \ref{fig:exp0:comp}A provides a summary of model-level differences between the three measures: probing accuracy (Acc Rep), MP Rep (Minimal Pair, representations), and MP Prob (Minimal Pair, probability). We observe that MP Rep underperforms MP Prob substantially in small models, but this disadvantage subsides and even reverses as model size increases (range: [-13.4\%, +6.89\%], correlation with log10(parameter count): $R = 0.791$, $p < 0.001$). These results suggest that in aggregate, probability-based comparisons tend to detect grammaticality contrasts more effectively than probing. However, these aggregate model-level trends conceal substantial variation across individual syntactic phenomena, which we examine next.

We find that there are BLiMP datasets that are in general (across models) better distinguished via probing over probability. If there is no particular pattern to when a given model underestimates grammatical knowledge via probability-based measures and vice versa, we expect a roughly uniform distribution of MP Rep > MP Prob over datasets. This is not the case; we find that dataset-level advantages exist along a continuum, with some datasets having stronger detectability via probing than via probability across nearly all models (labelled in Fig.\ \ref{fig:exp0:comp}B, left side), but the reverse is also true, with certain datasets having stronger detectability with probability-based measures (Fig.\ \ref{fig:exp0:comp}B, right side).

\subsection{Discussion}
This experiment establishes that mass-mean probing, and by extension the difference vector $\theta$, captures some key information about grammaticality, thus warranting further study. The LI-Adger dataset consists of sentences from an assemblage of grammatical contrasts, rather than one kind of grammatical violation. This suggests that our results are not simply due to some surface-level pattern or template-based similarities that are collinear with grammaticality. This is further supported by the fact that this probing setup tests for generalization to a different benchmark. In later experiments, this point is reinforced by further demonstrations of cross-phenomenon and cross-linguistic generalization. These results are consistent with the findings of \cite{wang2026ImplicitRepresentationsGrammaticality}, despite employing different probing techniques, sources of ungrammaticality, and representations to fit on.

We note that our probing accuracies are generally lower than \citeauthor{wang2026ImplicitRepresentationsGrammaticality}'s (\citeyear{wang2026ImplicitRepresentationsGrammaticality}) probes. This is unsurprising given the different goals of our experimental setup. We intentionally employ a low-complexity probe to evaluate how much grammaticality information is directly recoverable from the mass-mean difference vector, rather than maximizing classification performance. In addition, we do not tune the choice of representation layer for probing, and layer selection can substantially affect probe accuracy. Investigating how probe complexity and layer selection interact for mass-mean probing is a natural direction for future work, and may help clarify how much additional grammatical information can be extracted while preserving the interpretability of the probing method. The comparative results of this experiment also raise questions about how probing should be interpreted in tandem with results from probability-based approaches. We discuss this further in the Discussion section of this paper (\S\ref{sec:takeaway2}). Overall, we view the convergent results as preliminary evidence that the encoding of grammaticality is one of the major organizational principles of the representational space, at least in the larger language models ($\sim$1B+ parameters and above) that this work and the others examine.

\section{Experiment 2: Scope of grammaticality signals}
\label{sec:exp2}
Upon establishing that grammaticality distinctions are detectable via probing sentence representations, we would like to better understand the scope of the grammaticality signals. Specifically, we are interested in localization (\textit{where} the signals start to emerge) and cross-phenomenon generalization (how general the signal is beyond the specific phenomenon targeted).

For most ungrammatical sentences, there exists a point in the sentence where no grammatical continuation is possible. Do our probes capture grammaticality violations as the sentence unfolds, or do they only demonstrate sensitivity when the sentence is completed? We fit probes over intermediate representations extracted at the each of the sentence regions (Pre-critical, Critical, Final; \S \ref{sec:extract}) and evaluate them on held-out representations drawn from the same region. Based on our results from Experiment 1, we already expect to see sensitivity to grammaticality in the Final region; our main question is whether grammaticality distinction is captured in the Critical region as well. As control, we also report probing results in the Pre-critical region -- representations drawn from the token immediately preceding the Critical tokens. Since there is no grammaticality distinction in the Pre-critical region, we should expect to see no signal for probes fit in that region beyond noise.

We also explore the effects of phenomenon generality in grammaticality probing. In what ways are the grammaticality representations that we found specific or general, and how does this interact with sentential region? One hypothesis is that probes are highly-specific: they only succeed on datasets that are either the same dataset the probe was fitted on, or datasets of the same phenomenon.\footnote{The word ``phenomenon'' here refers to labels assigned to the datasets by us or the researchers that created the benchmarks (e.g., in the case of BLiMP). We acknowledge that the concept of phenomenon may not be present at all in a particular model's representation of sentences (though see \cite{kryvosheieva2025Differenttypessyntactic}), or they may not line up with the categorizations of phenomena assigned by us or by the creators of the benchmarks examined.} It predicts that probes would not generalize when tested on different phenomena. For this experiment, we fit a probe to one dataset, and evaluate that probe on other datasets or held-out sentences in the same dataset. We conduct this analysis on the 40 BLiMP subsets with Critical region annotations, yielding 1,600 (40 fit $\times$ 40 evaluation datasets) probes per model, per region.

The high-level summary of the results are as follows. First, we show that all models show probe sensitivity both at the Critical and Final regions, and not the control Pre-Critical region (\S\ref{sec:exp2:1}). The success profile of each of the models suggests that model representations of grammaticality are phenomenon-general. Nevertheless, we find that phenomenon-identity does influence probing accuracy -- probes fitted and tested on the same phenomenon are more accurate (\S\ref{sec:exp2:2}). Through a regression analysis, we show that various shared sentence properties explain variance in accuracy. Still, their effects are small and when regressed out, do not change the success profiles much -- the previously reported patterns of phenomenon-generality in probing success and phenonmenon-specificity in the success profiles remain (\S\ref{sec:exp2:3}).

\subsection{Result \#1: All models exhibit probe sensitivity
to grammaticality, from the moment grammaticality diverges}
\label{sec:exp2:1}
\begin{figure}[h!]
    \centering
    \includegraphics[width=\linewidth]{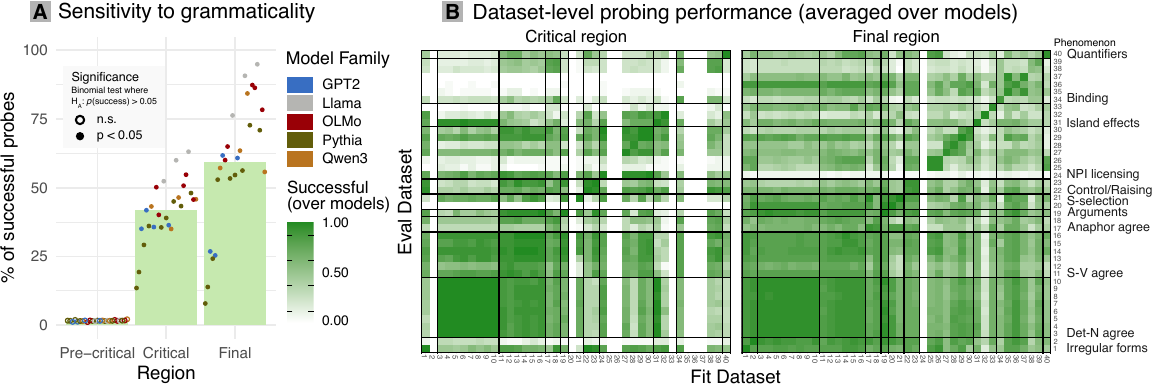}
    \caption{The success rate (as defined by \S\ref{methods:eval}) of mass-mean probes on detecting grammaticality across models (A) and across datasets (B). \textbf{(A)} Mean success rate across models in each region. Each dot represents a model's success rate, and dots are ordered by model scale. \textbf{(B)} Each grid represents mean success rate across models for a fit/evaluated dataset pair. Datasets are ordered by phenomena (labels on the right edge), and dataset labels are listed in Appendix \ref{app:blimp}.}
    \label{fig:exp1}
\end{figure}

\begin{figure}
    \centering
    \includegraphics[width=\linewidth]{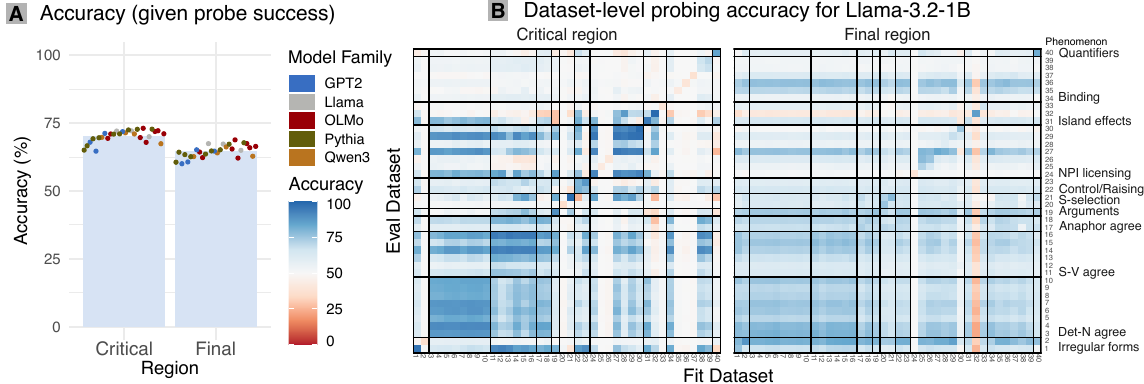}
    \caption{Probing accuracies of Experiment 2 in the Critical and Final region. (\textbf{A}) Mean accuracy rate across models in each region. (\textbf{B}) Accuracy profile for Llama-3.2-1B, a sample model with moderate performance relative to all models.}
    \label{fig:exp1acc}
\end{figure}
We find that all 25 models we examine demonstrate sensitivity to grammaticality in the Critical and Final regions and not the Pre-critical region. On average, the Final region has a higher proportion of successful probes than the Critical region (Fig.~\ref{fig:exp1}), though when probes succeed, probes at the Critical region were on average more accurate than a Final region probe (Fig.\ \ref{fig:exp1acc}). Overall, the results from this experiment are consistent with the findings from the previous experiment in a mixed-phenomena probing setup. Sensitivity at the Critical region also suggests that models maintain a grammaticality representation that is incrementally updated as the sentence unfolds. 

It is important to note that there is great variability in sensitivity among models, like what we have found in Experiment 1. In the Final region, the proportion of probe success ranged from 7.81\% (Pythia-14m) to 94.38\% (Llama-3.1-8B). At the model level, we find that model size (number of parameters) predicts a large swathe of this variance, accounting for 51.5\% variance in the Critical region and 59.0\% in Final region.

\subsection{Result \#2: Grammaticality signals generalize beyond dataset and phenomenon boundaries}
\label{sec:exp2:2}
If probes are phenomenon-specific, we expect that analyses that remove within-phenomenon probing results to yield little-to-no grammaticality signal or above-null probing performance. This is not the case. We only observe a minor drop in proportion of successful probes when we remove them from the analysis (Critical: average 4.06\% drop, [1.64\%, 5.28\%]; Final:  average 2.20\% drop, [-0.22\%, 5.68\%]). All models retained their significance with respect to sensitivity at both experimental regions. This is also captured visually via heatmaps (Fig.\ \ref{fig:exp1}B and Fig.\ \ref{fig:exp1acc}B): probe success is widely observed beyond the dataset/phenomenon diagonal, with the Final region yielding broader cross-phenomenon generalization than the Critical Region. These results show that the grammaticality signal captured by our probes is phenomenon-general, especially signals in the Final Region.

While probes, especially those in larger models are highly generalizable, it is also the case the probes that fit and test within the same phenomenon have a higher accuracy than its cross-phenomenon counterparts (Critical: average 14.0\% increase, [6.14\%, 18.26\%]; Final: average 6.64\% increase, [2.13\%, 11.03\%]). We find that phenomenon-specificity effects, as measured by the difference between cross- and within-phenomenon average probing accuracy are the strongest in small- to mid-size models (some examples are available Appendix \ref{app:models}, Fig.\ \ref{fig:app-heatmap}). In the next subsection, we explore whether surface-level properties of sentences, which may be correlated with phenomenon identity, explains the accuracy profiles observed in this experiment.

\subsection{Result \#3: Sentence similarity modulates accuracy, but does not affect probing success}
\label{sec:exp2:3}
Phenomenon-generalization could be capturing sentence-level similarity rather than abstract phenomenon-level representations; sentences targeting the same phenomenon, especially in the context of a grammaticality benchmark, will exhibit more surface-level similarities than sentences targeting a different phenomenon. Here, we ask whether similarity explains any part of probing accuracy. If it does, we ask whether the captured variance nullifies the probing successes we have reported.

For each probe accuracy matrix ($M \in \mathbb{R}^{40\times40}$, per region, per model), we compute its correlation to various measures of dataset similarity rooted in comparisons between sentence-level properties of the grammatical subset of the dataset. These properties are: length (KL-divergence of length distributions), lexical overlap (\% of common unique words), and representational similarity (average cosine similarity between sentence representations of the target region from the evaluated model). 

\begin{figure}
    \centering
    \includegraphics[width=\linewidth]{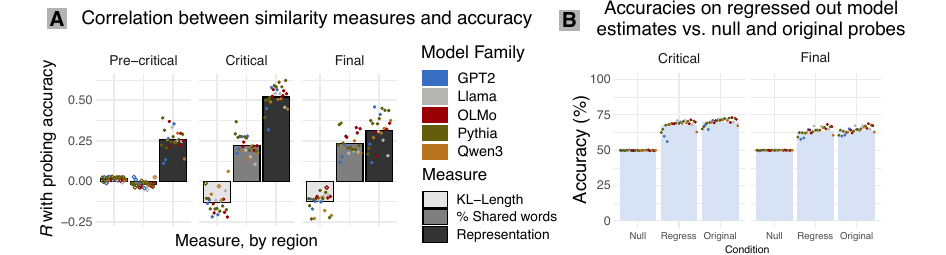}
    \caption{The relationship between surface-level similarity measures between BLiMP datasets and the probing accuracies reported in this Experiment. (\textbf{A}) The correlation between accuracies and similarity measures at each region. (\textbf{B}) Mean accuracies after regressing out the contribution of similarity.}
    \label{fig:exp1regress}
\end{figure}

Across models, we find that probing accuracies do correlate with various measures of between-dataset sentence similarity (Fig.~\ref{fig:exp1regress}A). They are also significant effects in a linear mixed-effects model, with the model structure \texttt{accuracy $\sim$ 1 + Length\_KL + Word\_Overlap + Rep\_Similarity + (1|model)}. The measures related to length and word overlap are correlated with probing accuracy in the Critical and Final regions, but not the Pre-critical region, whereas model representational similarity correlates with probing accuracy at all regions. However, LMEM estimates of accuracies without similarity terms (global intercept + random intercept + residual) remain substantially higher than the null distribution (Fig.~\ref{fig:exp1regress}B). This supports the claim that beyond sentence-level similarity, the grammaticality signals captured by mass-mean probing are phenomena general. The next experiments further aim to deconfound grammaticality from other likely correlated factors.

\section{Experiment 3: Deconfounding string probability from representations of grammaticality}
\label{section:exp3}
Could our probes be detecting a contrast of low vs.\ high probability strings rather than grammaticality? In all of the models we examine in this work, we know that the model assigns, on average, higher probability to a grammatical sentence over its ungrammatical counterpart (Appendix \ref{app:models}). While this difference is measured at the sentence pair level, we may expect a potential heuristic that achieves high accuracy on our grammaticality probing to be one that relies on probability as a cue rather than grammaticality. If the grammaticality signals detected by our probes were probability-specific rather than grammaticality-specific, then we expect a manipulation that decorrelates the two factors in the fitting sentence set (i.e., low probability grammatical vs. high probability ungrammatical) to yield at-chance predictions on grammaticality on held-out sentences. Alternatively, we expect minimal-to-no drop in accuracy if our probes were not sensitive to probability. This hypothesis also predicts that probability manipulations that strengthen this correlation will also have little-to-no effect on evaluation accuracy.

\begin{figure}[h]
    \centering
    \includegraphics[width=\linewidth]{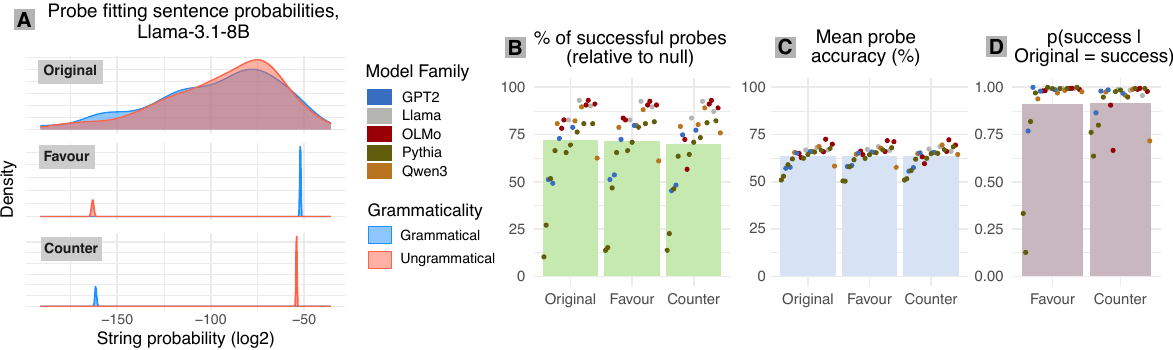}
    \caption{Results from Experiment 3. \textbf{(A)} The distribution of average fitting sentence probabilities for Llama-3.1-8B, across conditions. Each datapoint in the plot represents the mean sentence probability per fitting subset, separated by grammaticality label. \textbf{(B-D)} Probing performance between conditions: success, mean probe accuracy, and the rate of success given that the Original probe was successful.}
    \label{fig:exp3}
\end{figure}

This is the prediction we test: for each evaluated dataset, we created three probes corresponding to three probability-related manipulations. The Counter condition samples ungrammatical sentences from the 20\% most probable subset of sentences outside of the evaluated dataset, and grammatical sentences from the 20\% least probable sentences. The Favour condition reverses this by sampling grammatical sentences from the top 20\% probable subset and ungrammatical sentences from the least 20\% probable subset. Both of these conditions are compared against the Baseline condition, which does not impose any probability-related constraints on sampling. Fig.\ \ref{fig:exp3}A demonstrates the average string probabilities from the grammatical and ungrammatical subsets of each condition with Llama-3.1-8B as an example. We conduct these analyses on the Final representations of the BLiMP and SCaMP datasets. 

\subsection{Results}
We find that there are minimal differences of accuracy relative to the Original for both manipulation conditions. While a subset of models exhibit a significant drop in accuracy in the Counter condition, the magnitude was small: the largest accuracy drop across models was 3.19 percentage points (OLMo-2-0425-7B, Fig.\ \ref{fig:exp3}C). We also found similar degrees of negative accuracy modulations even in the Favour condition (Favour: [-2.61\%, +1.53\%] Counter: [-3.19\%, +6.12\%]). More importantly, we found that the Counter manipulation has nearly no negative effect on probe success for most models. This is demonstrated by measuring the proportion of Counter probe success given that its matching Original probe succeeds. We find that with the exception of the smallest scale models (left-hand side of Fig.\ \ref{fig:exp3}D), models are at-ceiling for both conditions.

Taken together, these results support the hypothesis that the probes are detecting grammaticality rather than string probability. These results do not contradict the hypothesis that model representations of grammaticality \textit{affect} string probability: an implicit assumption of the targeted syntactic evaluation paradigm. We discuss the relationship between our detection methods and string probability further in the Discussion section.

\section{Experiment 4: Deconfounding general sentence acceptability from grammaticality}
A prominent view in linguistics is that grammaticality and acceptability are linked but distinct \citep{schutze2016empiricalbaselinguistics,Schütze_Sprouse_2014}: grammaticality contributes to acceptability, which is measured through judgment tasks to draw inferences about grammaticality despite acceptability also reflecting other factors. This section asks whether our probes were detecting sentence acceptability instead of grammaticality. If this is the case, then we predict that the probes that are supposedly fitted to discern grammaticality will demonstrate sensitivity to other factors that modulate acceptability. We identify these factors based on previous human experimental work \citep{kluender1993Subjacencyprocessingphenomenon,hu2025ShadeszeroDistinguishing,tuckute2024Drivingsuppressinghuman,lau2017GrammaticalityAcceptabilityProbability} and utilize existing datasets to examine their detectability under a probe fitted on grammaticality contrasts. These factors include: semantic plausibility, event possibility/conceivability, truth, imageability, arousal, etc (Table \ref{app:dataset}). Throughout this experiment, we refer to such acceptability-modulating factors as ``extra-grammatical'' factors.

In this section, we present three analyses that test this overarching hypothesis. The first two analyses seek to understand whether there is any effect of acceptability-modulating extra-grammatical factors on our grammaticality probes. Finding that there is an effect, we inquire, through a probe boundary-tilting experiment, whether acceptability subsumes grammaticality, or whether there remains a distinct encoding for grammaticality. Overall, we find that linear probes fitted on grammaticality contrasts are a composite of both grammaticality and acceptability signals, with the former still being the dominant signal.

\subsection{Shared encoding between extra-grammatical contrasts and grammaticality}
We can further examine the representational relationship between various acceptability-related contrasts and grammaticality by testing how encodings for other contrasts generalize to grammaticality datasets, and vice versa. The strong hypothesis that our grammaticality probes have no signal of extra-grammatical factors, will predict that probes fitted on an extra-grammatical dataset should not generalize at all to grammaticality datasets. In contrast, if there exists some general sentence acceptability encoding, then generalization between these datasets are expected. The setup of this experiment was similar to Experiment 1: we fitted a probe for each extra-grammatical dataset (see Table \ref{tab:extragram-dat} for benchmark/dataset information) or BLiMP dataset, and evaluated the probe on each BLiMP dataset or extra-grammatical dataset, respectively.

\begin{figure}[h]
    \centering
    \includegraphics[width=\linewidth]{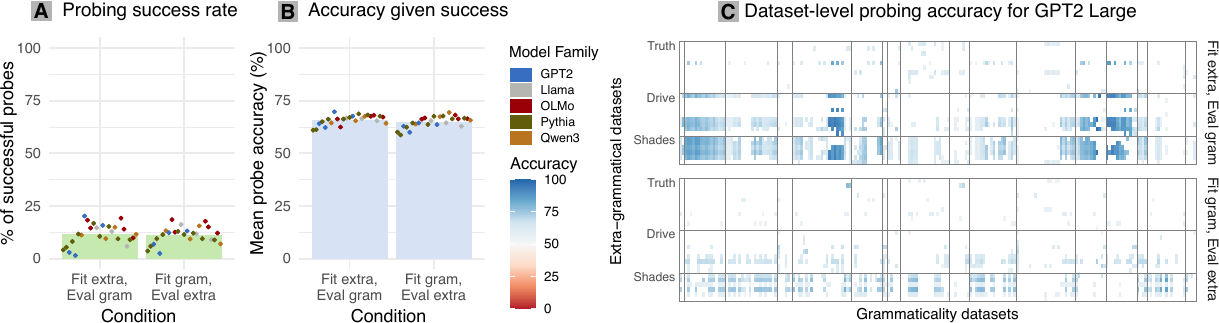}
    \caption{Probing results when grammaticality probes are tested on extra-grammatical datasets and vice versa. \textbf{(A)} Success rates of probes across models. \textbf{(B)} Mean probe accuracy given that a probe is successful. \textbf{(C)} Generalization profile for GPT2 Large (model with strongest grammatical <-> extra-grammatical probe 
    generalization) as an example. Vertical lines indicate grammaticality phenomenon boundaries, horizontal lines indicate extra-grammatical benchmark boundaries.}
    \label{fig:exp4cross}
\end{figure}

Across models, we observe some degree of generalization of grammaticality probes to extra-grammatical sentence representations and vice versa (Fig.\ \ref{fig:exp4cross}A). Though the percentage of probes that demonstrate generalization are far lower than reported in previous experiments (gram $\to$ extra mean: 11.2\%, extra $\to$ gram mean: 11.5\% vs. gram $\to$ gram mean: 55.9\%, Experiment 2), it is still a non-negligible degree of generalization that merits attention. In particular, we find that this generalization is observed for a few select extra-grammatical datasets, and near-to-no sensitivity in others. Fig.\ \ref{fig:exp4cross}C illustrates this selective extra-grammatical sensitivity for GPT2 Large, the model with the strongest generalization between grammaticality and extra-grammaticality probes.

\subsection{Minor probing accuracy differences between SCaMP-Plausible vs.\ SCaMP-Implausible}
The SCaMP dataset offers us a window into measuring the effect of semantic plausibility on probing performance. Each SCaMP dataset has plausible and implausible variants; all sentences in the implausible variant, regardless of grammaticality label, violate selectional category restrictions on verbs. Ungrammatical sentences in the plausible datasets and all sentences from the implausible dataset should fall under the same category from the perspective of an acceptability-tracking probe, as both are anomalous, albeit for different reasons. Therefore, if it is the case that our grammaticality probes are acceptability-tracking rather than grammaticality-tracking, we expect a probe fitted to differentiate grammaticality on a plausible dataset to poorly differentiate the grammaticality of an implausible dataset. This is the prediction we test, employing the same probe-fitting procedures as Experiment 2. To quantify poor differentiation in implausible datasets, we make comparisons to how these probes perform on matching plausible datasets. If it is the case that a probe poorly differentiates grammaticality due to general sentence anomaly, then we expect a drop in accuracy on the implausible dataset when compared against its plausible counterpart.

\begin{figure}[h]
    \centering
    \includegraphics[width=\linewidth]{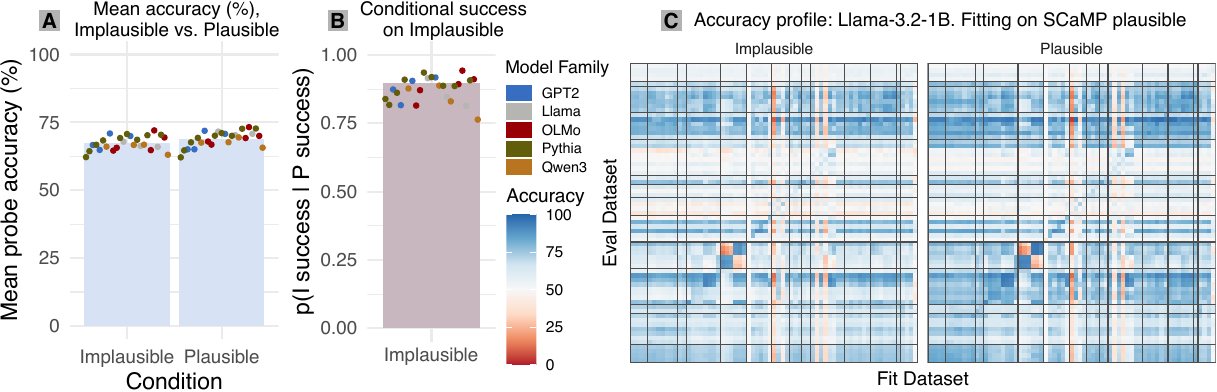}
    \caption{Probing results on SCaMP plausible and implausible datasets. The fitting datasets are always in the plausible subset but the evaluated dataset are either plausible are implausible \textbf{(A)} Mean probing accuracy between plausibility conditions. \textbf{(B)} Given success on a (Fit$_\text{Plausible}$, Eval$_\text{Plausible}$) probing configuration, the rate of success for matching Eval$_\text{Implausible}$ datasets. \textbf{(C)} Accuracy profile for Llama-3.2-1B, as an example. Lines on the heatmap indicate phenomenon boundaries.}
    \label{fig:exp4scamp}
\end{figure}

We find that there are measurable accuracy and success differences based on the plausibility of the evaluated dataset, though this effect is minor. Across models, the mean accuracy drop was 1.06 percentage points, with the range of within-model mean accuracy differences (Acc(Plausible) $-$ Acc(Implausible)) being [-0.57\%, 4.34\%] (Fig.\ \ref{fig:exp4scamp}A). The conditional success rate is relatively high (mean: 90.2\%, range: [80.3\%, 95.8\%]), but not at-ceiling (Fig.\ \ref{fig:exp4scamp}B). At the dataset level, the accuracy profiles between plausible and implausible datasets are globally similar, but with the implausible datasets having a weaker signal than that of their plausible counterparts. Fig.\ \ref{fig:exp4scamp}C provides an example of this for a specific model (Llama-3.2-1B), and this relationship is quantified by measuring the correlation between the two accuracy matrices -- in Fig.\ \ref{fig:exp4scamp}C, $R = 0.927$. Across models, the correlation ranges from 0.763 to 0.964. Overall, while sentence plausibility has a measurable effect on our grammaticality probes, the effect is modest, indicating that grammaticality remains a robust and reliable signal across both plausible and implausible sentences.

\subsection{Factoring out extra-grammatical acceptability signals with boundary tilting}
The results from the previous analysis suggest that mass-mean probes capture some degree of general sentence acceptability, but leaves unclear the degree to which this signal subsumes the grammaticality signal reported in the previous experiments. In this experiment, we first assume that the acceptability signals can be factored out of the probing boundary. Taking that assumption further, if it is the case that acceptability subsumes grammaticality, we predict that probes with their acceptability signals factored out would render them poor classifiers of our grammaticality datasets.

We remove acceptability signals via boundary tilting, regressing out the shared signal among the extra-grammatical and grammatical datasets from the probing difference vector $\theta$. This regression is done by first computing the vector $\phi$: the difference vector for an extra-grammatical dataset. Then, we remove the shared variance between $\theta$ and $\phi$ by subtracting the vector projection of $\theta$ onto $\phi$ ($\frac{\theta\cdot\phi}{\|\phi\|}\,\frac{\phi}{\|\phi\|}$) from $\theta$, generating the tilted vector $\theta_\phi$. $\theta_\phi$ is then transformed into a mass-mean probe $\sigma(\theta_\phi \cdot x)$ for evaluation.

We compare between conditions by comparing the probing accuracies or success over the same evaluation datasets. The conditions vary in terms of what $\phi$ is. The Tilt-all condition takes the mean the all difference vectors from each of the extra-grammatical datasets as $\phi$. However, failure to see a reduction in probing performance could stem from the mean operation, which may thin out the acceptability signal that can be present in some difference vectors but not others. Thus, we also define a Tilt-one condition where we compute $\phi$ from the extra-grammatical dataset with the highest cosine similarity between $\theta$ and $\phi$. Each evaluated dataset has a probe (across 20 runs) fitted for each condition. A further Control condition used to validate the boundary tilting method defines $\phi$ by randomly sampling 26 difference vectors from other grammaticality probes (for other evaluated datasets) and taking the mean, akin to the Tilt-all condition. If the method works, then we expect that the Control condition to render probes that are classifying at-chance.

\begin{figure}[h]
    \centering
    \includegraphics[width=\linewidth]{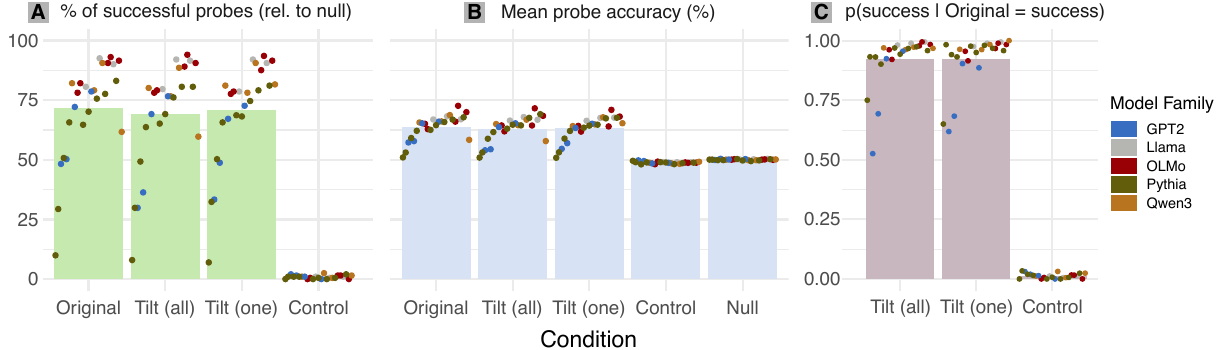}
    \caption{Probing results across different $\theta$, pre- and post- tilting. All probes are evaluated on grammaticality datasets (BLiMP and SCaMP). \textbf{(A)} Success rates across conditions. \textbf{(B)} Accuracy scores across conditions. \textbf{(C)} Success rates of tilted and control probes, given success of the original probe.}
    \label{fig:exp4}
\end{figure}

The results indicate that grammaticality is a distinct signal from acceptability. First, we see that the equivalent tilting over grammatical datasets (Control) causes probes to be unable to distinguish grammaticality, whereas the Tilt conditions essentially pattern the same as the Original probe. These results are consistent among three reporting measures: success (Fig.\ \ref{fig:exp4}A), accuracy (Fig.\ \ref{fig:exp4}B), and success conditioned on success on the matched Original probe (Fig.\ \ref{fig:exp4}C). Like Experiment 3, we find that the models that are most affected by this manipulation are smaller models.

\section{Experiment 5: Cross-linguistic generalization of grammaticality}
We now turn to studying the language-generality of our grammaticality probes. Large language models are often explicitly trained on multiple languages and can respond to queries from non-English languages (\citeauthor{behzad2024AskLLMsEnglish}, \citeyear{behzad2024AskLLMsEnglish}; though see \citet{ahuja2023MEGAMultilingualEvaluation} for notable performance gaps between languages). A key assumption in generative linguistics is that, despite extensive typological variation, the core machinery underlying grammaticality distinctions is shared across languages \citep{chomsky1995MinimalistProgram}. Is the encoding of grammaticality in the NLMs that we study also language general?

This experiment focuses on the five models that exhibit the strongest grammaticality signals from the previous experiments, both in terms of probe success rate and accuracy: Llama-3.2-3B, Llama-3.1-8B, Qwen3-4B, OLMo-7B-hf, and OLMo-2-1124-7B. As a precondition to studying the language-generality of grammaticality encodings, we first establish that for these models, there is above-chance probing signal in the non-English languages we examine. Then, we establish that language-general encodings of grammaticality exist with a cross-linguistic probing analysis. We conduct our analyses on six benchmarks (CLAMS-\{de, fr, ru, he\}, RUBLiMP, and ZhoBLiMP; see Table \ref{tab:gram-dat}) across five languages: German, French, Russian, Hebrew, and Mandarin Chinese. 

The probe-fitting and evaluation procedures closely follow Experiment 2 (\S\ref{sec:exp2}). In the within-language analysis, we fit probes on sentences sampled from same-language datasets that are not the evaluated dataset. In the cross-linguistic analysis, we fit probes on sentences sampled from BLiMP and SCaMP and evaluated them on the non-English datasets.

\subsection{Result \#1: Successful within-language grammaticality detection for a variety of non-English languages}
\label{sec:crossling-within}
\begin{figure}[h]
    \centering
    \includegraphics[width=\linewidth]{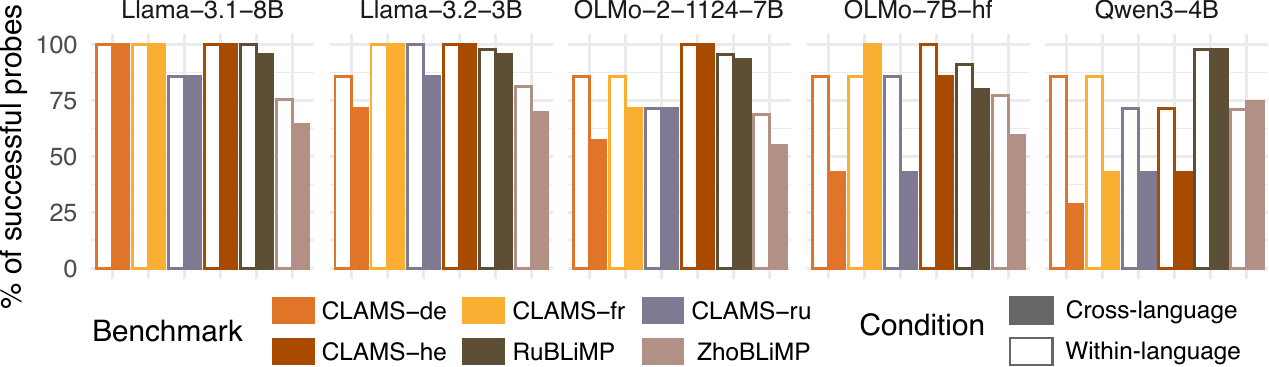}
    \caption{Within-language (left bars) and cross-language (right bars) probing success rates across the five models examined. Each bar represents one benchmark, each on a language. (Note: RuBLiMP and CLAMS-ru are both Russian benchmarks)}
    \label{fig:exp5:within}
\end{figure}
We find that internal grammaticality distinctions are also present for non-English languages (Fig.\ \ref{fig:exp5:within}, left bars). This is the case even for language models that are not explicitly trained on datasets from these language, as indicated by results from OLMo \citep{olmo20252OLMo2}. Some model-benchmark pairs achieve near-ceiling success rates, such as Llama models on Russian benchmarks. The accuracy profiles of these successful probes are comparable to English (English mean across the five models: 66.6\%, [65.0\%, 68.8\%]; Cross-ling mean: 61.4\%, [59.9\%, 62.9\%]). These results indicate that the models examined have non-trivial grammaticality distinctions for various languages other than English.

\subsection{Result \#2: Shared grammaticality encoding across languages in some models}
Now, we ask whether probes fitted on languages other than the evaluated dataset's language yield similar degrees of success and accuracy. If the encoding of grammaticality is language-general, then we expect there to be little difference on both success and accuracy, regardless of whether the probe was fitted on the same language or a different language.

As can be seen from Fig.~\ref{fig:exp5:within} (right bars), we find that there is a decrease in successful probes in the cross-linguistic condition across all five models, albeit to substantially different degrees in different models. We find that the drop in success rates in the Llama models is overall marginal, whereas there is a drastic drop in success rates in OLMo-7B-hf and Qwen3-4B. However, there still exists a non-trivial number of successful cross-linguistic probes in all models. We take this analysis a step further and analyze the probing performance for all the language pairs examined in this experiment. 

\begin{figure}[h]
    \centering
    \includegraphics[width=\linewidth]{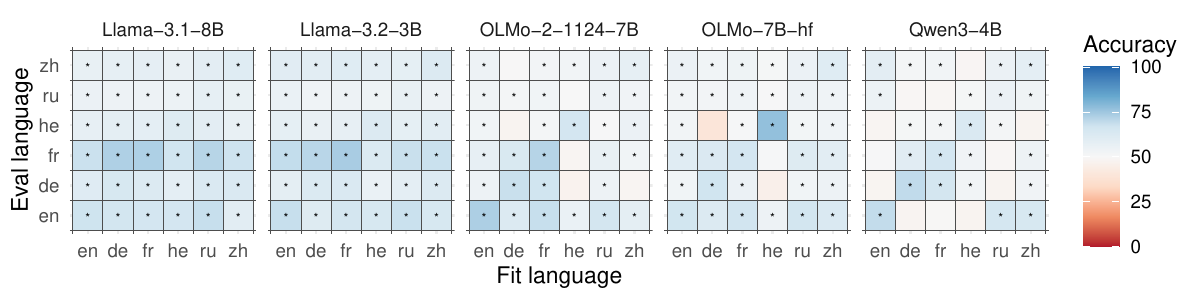}
    \caption{Probing results across six languages, across five high-performant models. ``*'' marked in a grid indicates that the probe for that language pair was successful.}
    \label{fig:exp5}
\end{figure}

Consistent with the previous analysis, we find that there is detectable grammaticality signal across many language pairs, though generalizability between pairs is not guaranteed in all models (Fig.\ \ref{fig:exp5}). We note that in this analysis, we collapse the distinctions between phenomena within a language by aggregating them for evaluation, which obscures the interesting accuracy variation that may help us diagnose why a language pair does not generalize for a given model. Overall, the results from this experiment suggest that NLMs can develop an encoding of grammaticality that is language-general, and does so with high fidelity in models explicitly trained on multilingual text \citep[Llama models:][]{grattafiori2024Llama3Herd}. This once again provides evidence that grammaticality is a major organizational dimension in the representational space, at least in some language models. There are many follow-up questions about the development of a consistent, language-general representation of grammaticality, including when such representations emerge during training, how they relate to language-specific syntactic features, and whether they depend on shared structural regularities across languages or instead arise from the pressure to model multilingual data efficiently.

\section{General Discussion}
Our series of experiments find that there is a reliably detectable representational separation between grammatical and ungrammatical linguistic expressions in the models we examine. This representational divide is captured by a single linear direction, and this direction is highly generalizable across phenomena and languages in stronger models. We discuss the implications of these findings and highlight future connections that can be made to better understand model grammaticality and syntactic knowledge.

\subsection{RQ1: Detection of grammaticality from sentence representations}
Through this work, we find that a low-complexity probe over sentence representations is able to reliably predict grammaticality on held-out sentences. Our findings are consistent with \cite{wang2026ImplicitRepresentationsGrammaticality} in that regard, except the probing parameter $\theta$ can be readily manipulated/tilted, as seen in our later experiments. For models with consistently high probing scores, these empirical results suggest that for an arbitrary linguistic expression, the models' representation of this expression contains an abstract encoding of whether it is grammatical or not. The findings from Experiment 2 demonstrate that grammaticality is continually represented as the sentence unfolds. Future work could connect this incremental property of grammaticality representations to previous work detecting parse states \citep{eisape2022ProbingIncrementalParse} and dependencies as the sentence unfolds \citep{hewitt2019StructuralProbeFinding}.

Zooming out from these particular findings, it is important to ask: why do we see linear representational separation for grammaticality at all? All of the models we examine were trained on an autoregressive objective, and information about sentence grammaticality is generally not explicitly labelled in the pretraining data. \cite{ravfogel2025EmergenceLinearTruth} develop a hypothesis for why we may expect to see linear separation for truth \citep{burns2022DiscoveringLatentKnowledge,azaria2023InternalStateLLM,marks2024GeometryTruthEmergent}: factual statements tend to co-occur with each other while falsehoods co-occur with other falsehoods. An autoregressive objective encourages the model to learn this distinction (of which statements are factual, and the correlation over truth/false properties) to minimize loss. It is unclear to us yet whether grammaticality as a property has similar co-occurrence structure in training data -- this is a testable question in the models we investigate with their training data publicly available (e.g., the Pythia and OLMo series). In future work, one may apply training data attribution methods to also finely examine what particular properties of training utterances and their statistics lead to the emergence of grammaticality encodings. Another related question that should be addressed is whether representational separation for grammaticality is inherently tied to a model's ability to represent syntactic knowledge. If that is the case, then explanations about why, or when, we expect to find linear encodings of grammaticality can be tied to understanding the conditions that give rise to syntactic representations. Finally, there exists utility-based explanations of the development of a grammaticality encoding. While these models are not the models that are deployed as chatbots in the current age, its training data may consist of conversations or documents that involve metalinguistic information. This consists of papers about grammaticality (e.g., the BLiMP paper itself), or language instructional conversations. In our selection of models to examine in this work, we have avoided the use of chat-based models for this reason, but this information may still be present in the pretraining data. In the future, if we were to study the the development of syntactic representations alongside grammaticality representations in relation to humans, we would need to conduct the same analyses presented in this work on a model that is trained on carefully vetted data that control for the amount and type of metalinguistic information.

\subsubsection{Syntactic representations and grammaticality representations}
In the study of human language, syntactic representations and grammaticality are tightly linked: on generative accounts, ungrammaticality often signals the lack of a licit syntactic structure for a given expression  \citep{chomskyAspectsTheorySyntax}.Our probes establish the existence of a grammaticality boundary, but a similar link with syntactic representations has yet to be established. Furthermore, the mechanism through which an expression is placed into the grammatical or ungrammatical region remains unknown. Future work may study whether sentences in the ungrammaticality region lack representational signatures for structure (e.g., probe failures for constituency or dependency structures: \citealt{tenney2019what}), and eventually establish causal links between structural representations and the grammaticality boundary we identify.

Through our experiments, we also find that there exist specific types of grammatical contrasts that are not detectable across models. In general, it can also be asked for any specific model studied in this work why contrasts tied to specific phenomena are undetectable while maintaining robust representations of grammaticality in general. It could be the case that this representational distinction exists internally but is undetectable via mass-mean probing, but then the question of why not, for these specific phenomena, still remains. Future work may test whether characteristics specific to the phenomena in question, especially distributional properties in the training data, predict detectability in representation space.

\subsection{RQ2: Distinctions between grammaticality and collinear properties of linguistic expressions}
\label{sec:takeaway2}
As discussed in the Introduction, a persistent argument against the claim that NLMs have grammaticality distinctions has been that they fundamentally entangle sentence likelihood and grammaticality. Through Experiment 3, we demonstrate that this is in fact not the case; we find that there is little-to-no reduction in probing performance when we decouple string probability and grammaticality, showing that the grammaticality probes reported in our experiments are not merely sensitive to string probability. In investigating the distinction between grammaticality and acceptability in Experiment 4, we have also developed methods for decoupling the many collinear factors that may exist in the encoding of a sentence representation, results from which further support the existence of a grammaticality boundary independent from those factors. We recommend using similar methods to deconfound these factors in representation space to complement probability-based measures of grammaticality. 

These findings also have practical implications for researchers studying grammaticality in NLMs. In Experiment 1, we report that probability-based and probe-based methods captured different sources of variation, suggesting that they should be viewed as complementary rather than interchangeable measures. In particular, the additional variance explained by representation-based probes indicates that grammatical information encoded in sentence representations does not necessarily propagate into the model's output probability distribution. Consequently, relying exclusively on probability-based measures may underestimate the extent to which grammatical distinctions are represented internally. One important direction for future work is to characterize the relationship between representational and probabilistic measures of grammaticality, including why the two capture systematically different sources of variation. Together with questions about how these findings relate to models' syntactic representations \citep{hewitt2019StructuralProbeFinding,tenney2019BERTRediscoversClassical}, answering these questions may clarify what aspects of grammatical knowledge are reflected in pretrained LM probabilities, with important implications for their use in computational psycholinguistics.

\subsection{RQ3: The scope of grammaticality representations}
We find that representations of grammaticality, whenever they are detectable for a given model and a given syntactic phenomenon, the identified probe boundary is often generalizable to other phenomena. We find that the phenomenon-general nature of grammaticality is strengthened as model size increases. In Experiment 5, we find that for a select set of models, their grammaticality encoding is language-general. These generality results further reinforce the interpretation that the representational separation is selective to grammaticality, rather than to low-level features of a minimal pair dataset correlated with grammaticality.

Connecting our findings to the broader literature, we see that grammaticality distinctions are not limited to distinguishing ``naturally occurring'' or frequently observed ungrammaticality. In our work, the distinctions were observable for datasets that contain ungrammatical contrasts that are commonly observed in speech errors \citep[e.g., subject-verb agreement errors;][]{fromkin1971NonAnomalousNatureAnomalous,bock1991Brokenagreement} and also for datasets with word shuffles that are less frequently observed but nonetheless elicit robust ungrammaticality judgments from humans. Results from \citeauthor{wang2026ImplicitRepresentationsGrammaticality}'s (2026) probes suggest that the grammaticality distinctions that we find are likely extendable to even more extreme cases of unnatural sentence perturbations. In future work, we plan to further isolate grammaticality by studying in what ways the grammaticality encodings reported in our study persist under various nonce-word (``jabberwocky'') manipulations to both ungrammatical and grammatical sentences.

\subsection{On the utility of NLMs for human cognitive science}
How do the results of this study inform larger questions related to the use of NLMs for the study of human language and cognition? As outlined in the introduction, sensitivity to grammaticality distinctions is a requisite premise for the view that NLMs can serve as models of human language, but whether this premise holds in the first place is debated. We view our work as settling this premise, thus opening up future avenues for arguing precisely, in what ways, the study of the linguistic capabilities of NLMs can contribute to the study of the human language faculty. The results suggest that the central question is no longer whether NLMs can represent grammaticality distinctions at all, but rather which distinctions they acquire, how robustly they acquire them, and under what architectural and learning conditions these distinctions emerge. By shifting the debate in this direction, our findings provide a clearer foundation for investigating the relationship between grammatical competence in artificial and human learners.

To conclude, the main contributions of this work are twofold. First, we provide a representation-based method complementary to existing probability-based methods for measuring a model's ability to distinguish grammaticality for a given grammatical phenomenon. Second, in doing so, we demonstrate that it is indeed possible for statistical systems, in this case neural network learners of language, to exhibit robust and generalizable grammaticality distinctions. Future studies can build on this method to investigate the architectural and data conditions under which grammaticality distinctions arise, as well as the extent to which such distinctions generalize across a wider range of linguistic phenomena, languages, and and model classes. Regardless of one's belief about whether these models can contribute directly to the study of human cognitive science, our results establish that the emergence of grammaticality distinctions is not unique to symbolic or explicitly rule-based systems, and therefore constitutes an empirical phenomenon that demands explanation in its own right.

\section*{Acknowledgments}
This work is supported by a SSHRC Doctoral Fellowship (752-2024-0291) to JL. We acknowledge that the computational work reported on in this paper was performed on the Shared Computing Cluster which is administered by \href{https://www.bu.edu/tech/support/research/}{Boston University's Research Computing Services}. We thank tinlab at Boston University, the Group for Language and Intelligence and the Semantics Lab at Johns Hopkins University, and audiences at the GenLM reading group for their feedback on this work.

\bibliographystyle{apalike}  
\bibliography{jl.bib}  
\newpage
\begin{appendices}
\input{app-methods.tex}

\section{Experiment 2}
\subsection{Additional accuracy profiles}
\begin{figure}[h!]
    \centering
    \includegraphics[width=0.8\linewidth]{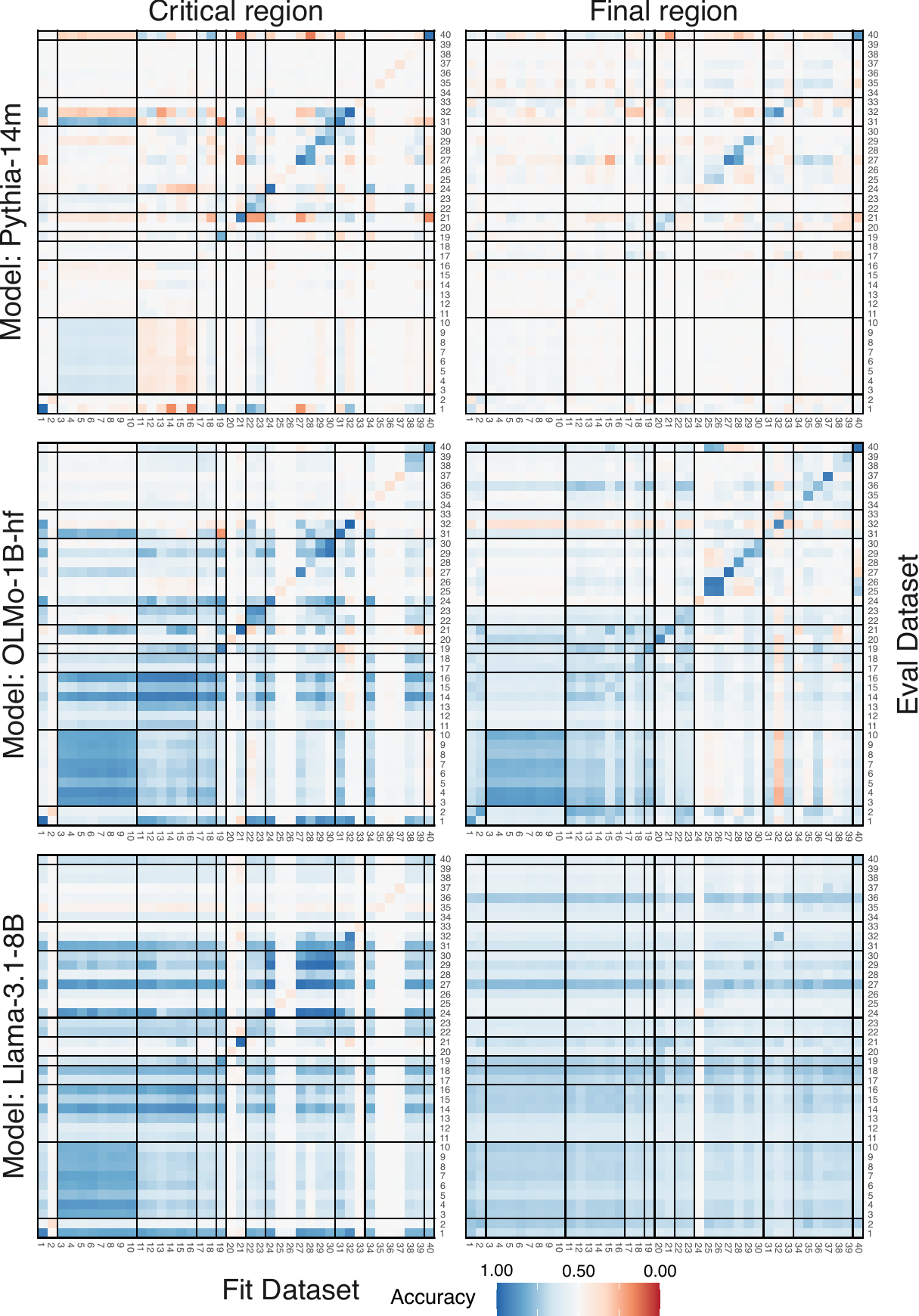}
    \caption{Critical and Final region accuracies on BLiMP dataset-level probing (Experiemnt 2) on Pythia-14m, OLMo-1B-hf, and Llama-3.1-8B.}
    \label{fig:app-heatmap}
\end{figure}

We report additional accuracy profiles from models in Fig.\ \ref{fig:app-heatmap} across the success spectrum to better understand how phenomenon-generality differ across models and regions.

\end{appendices}

\end{document}

%% file: app-methods.tex
\section{Methods}
\subsection{Dataset preprocessing}\label{app:dataset}
\paragraph{SCaMP}
Datasets from SCaMP are not labelled for the syntactic phenomenon they belong to. We manually labelled the syntactic phenomenon of each dataset, with most datasets falling into phenomenon categories set up by the authors of BLiMP. In addition to the BLiMP phenomena, we categorized some SCaMP datasets into novel categories. These categories and their associated datasets are documented in Table \ref{tab:scamp-cat}.

\begin{table}[h!]
    \centering
    \begin{tabular}{lp{10cm}}
    \toprule
        Phenomenon & Datasets \\
    \midrule
        Adverb/Adj restriction & \texttt{recursion{\textunderscore}intensifier{\textunderscore}\{adj,adv\}{\textunderscore}\{short,medium,long\} {\textunderscore}\{plausible,implausible\}}\\
        Center embedding & \texttt{center{\textunderscore}embedding{\textunderscore}\{single, double\}{\textunderscore}\{1,2\} {\textunderscore}\{plausible,implausible\}}\\
        Verb count & \texttt{recursion{\textunderscore}pp{\textunderscore}\{is,verb\}{\textunderscore}\{short,medium,long\} {\textunderscore}\{plausible,implausible\}}\\
        Word order & \texttt{swapped{\textunderscore}ditransitive{\textunderscore}\{1,2\}{\textunderscore}\{plausible,implausible\}, \{pp,adv\}{\textunderscore}order{\textunderscore}\{plausible,implausible\}, svo{\textunderscore}vos{\textunderscore}\{plausible,implausible\}, green{\textunderscore}ideas{\textunderscore}\{plausible,implausible\}}\\
    \bottomrule\\
    \end{tabular}
    \caption{Novel phenomenon categories and associated datasets for SCaMP. The remainder of datasets were categorized following BLiMP's categorization scheme.}
    \label{tab:scamp-cat}
\end{table}

\paragraph{DGL}
The DGL dataset contains sentences from \cite{dentella2023Systematictestingthree} as well as minimal pair counterpart sentences created by \cite{hu2024Languagemodelsalign}.

\paragraph{Shades}
The Shades benchmark consists of sentences derived from the multi-condition minimal pairs stimulus set from \cite{hu2025ShadeszeroDistinguishing}. Each sentence from the stimulus set consists of \texttt{probable}, \texttt{improbable}, \texttt{impossible}, \texttt{inconceivable}, and \texttt{inconceivable{\textunderscore}syntactic} continuations at the critical word. In particular, we selected the first four conditions (\texttt{inconceivable{\textunderscore}syntactic} involves a grammaticality contrast), and created minimal pair datasets from each possible pair of conditions. This yields 4C2 = 6 datasets.

\paragraph{Drive}
The Drive benchmark consists of sentences derived from the stimulus set of \cite{tuckute2024Drivingsuppressinghuman}. Each sentence in the stimulus set is rated for the following measures: \texttt{grammaticality}, \texttt{arousal}, \texttt{conversational}, \texttt{sense}, \texttt{frequency}, \texttt{imageability}, \texttt{others{\textunderscore}thoughts}, \texttt{physical}, \texttt{places}, and \texttt{valence}. Since we are interested in extragrammatical factors, we filter the stimulus set for sentences that are rated above 4.0 on grammaticality on a Likert scale (average, across raters), rendering 1542 sentences. For each of the extragrammatical dimension, we compute the rating mean for that dimension, and label the filtered sentences with the mean as a threshold.

\paragraph{Truth}
We selected the following datasets from \cite{marks2024GeometryTruthEmergent}: \texttt{sp{\textunderscore}en{\textunderscore}trans}, \texttt{neg{\textunderscore}sp{\textunderscore}en{\textunderscore}trans}, \texttt{common{\textunderscore}claim{\textunderscore}true{\textunderscore}false}, \texttt{cities}, \texttt{neg{\textunderscore}cities}, \texttt{cities{\textunderscore}cities{\textunderscore}conj}, \texttt{cities{\textunderscore}cities{\textunderscore}disj}, \texttt{larger{\textunderscore}than}, \texttt{smaller{\textunderscore}than}, \texttt{counterfact{\textunderscore}true{\textunderscore}false}, and \texttt{companies{\textunderscore}true{\textunderscore}false}.

\subsection{BLiMP dataset labels}\label{app:blimp}
In heatmaps referencing BLiMP datasets, we used numerical labels to make space. Table \ref{tab:blimp-labels} maps these numerical labels to each of the BLiMP datasets, along with their phenomenon group ascribed by BLiMP \citep{warstadt2020BLiMPBenchmarkLinguistic}.
\input{blimp.tex}

\subsection{Models}\label{app:models}
\begin{figure}[h!]
    \centering
    \includegraphics[width=0.8\linewidth]{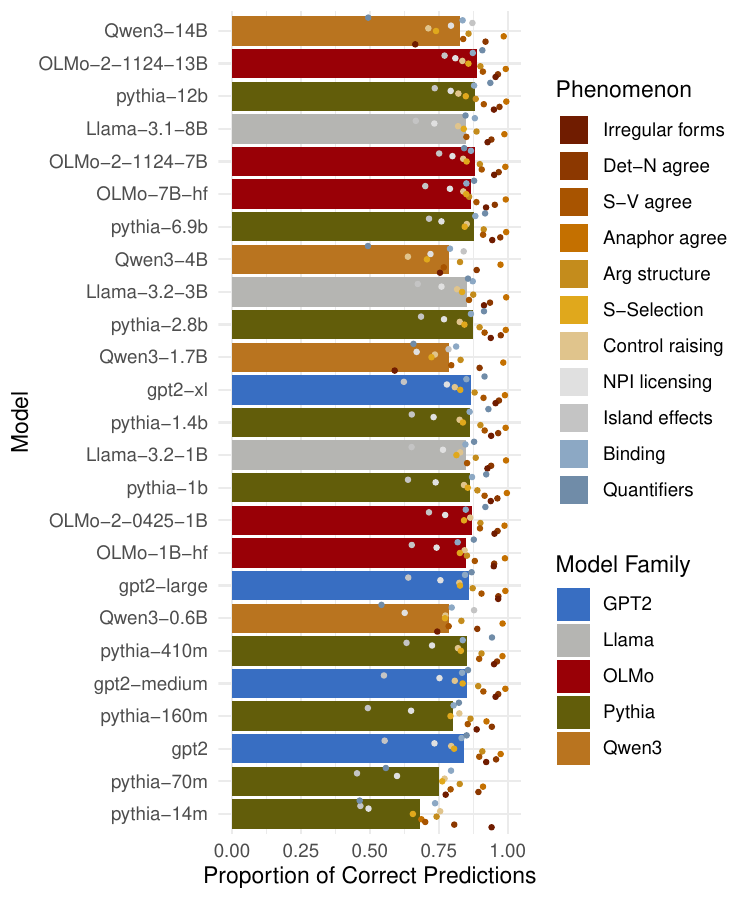}
    \caption{Minimal pair probability comparison-based BLiMP accuracies for each model we examine. Each dot on the bar represents the model's accuracy on a particular phenomenon. Models are organized from largest (in number of parameters; top) to smallest.}
    \label{fig:appprob}
\end{figure}

Fig.\ \ref{fig:appprob} summarizes model performance with the targeted syntactic evaluation paradigm on the BLiMP datasets, broken down by phenomenon.

%% file: blimp.tex
\begin{table}[h!]
\centering

\begin{tabular}{cll}
\toprule
\# & Dataset & Phenomenon\\
\midrule
1 & \texttt{anaphor{\textunderscore}gender{\textunderscore}agreement} & Anaphor agreement \\
2 & \texttt{anaphor{\textunderscore}number{\textunderscore}agreement} & Anaphor agreement \\
3 & \texttt{determiner{\textunderscore}noun{\textunderscore}agreement{\textunderscore}1} & Det-N agree \\
4 & \texttt{determiner{\textunderscore}noun{\textunderscore}agreement{\textunderscore}2} & Det-N agree \\
5 & \texttt{determiner{\textunderscore}noun{\textunderscore}agreement{\textunderscore}irregular{\textunderscore}1} & Det-N agree \\
6 & \texttt{determiner{\textunderscore}noun{\textunderscore}agreement{\textunderscore}irregular{\textunderscore}2} & Det-N agree \\
7 & \texttt{determiner{\textunderscore}noun{\textunderscore}agreement{\textunderscore}with{\textunderscore}adj{\textunderscore}2} & Det-N agree \\
8 & \texttt{determiner{\textunderscore}noun{\textunderscore}agreement{\textunderscore}with{\textunderscore}adj{\textunderscore}irregular{\textunderscore}1} & Det-N agree \\
9 & \texttt{determiner{\textunderscore}noun{\textunderscore}agreement{\textunderscore}with{\textunderscore}adj{\textunderscore}irregular{\textunderscore}2} & Det-N agree \\
10 & \texttt{determiner{\textunderscore}noun{\textunderscore}agreement{\textunderscore}with{\textunderscore}adjective{\textunderscore}1} & Det-N agree \\
11 & \texttt{irregular{\textunderscore}past{\textunderscore}participle{\textunderscore}adjectives} & Irregular forms  \\
12 & \texttt{irregular{\textunderscore}past{\textunderscore}participle{\textunderscore}verbs} & Irregular forms \\
13 & \texttt{distractor{\textunderscore}agreement{\textunderscore}relational{\textunderscore}noun} & Subj-V agree \\
14 & \texttt{distractor{\textunderscore}agreement{\textunderscore}relative{\textunderscore}clause} & Subj-V agree \\
15 & \texttt{irregular{\textunderscore}plural{\textunderscore}subject{\textunderscore}verb{\textunderscore}agreement{\textunderscore}1} & Subj-V agree\\
16 & \texttt{irregular{\textunderscore}plural{\textunderscore}subject{\textunderscore}verb{\textunderscore}agreement{\textunderscore}2} & Subj-V agree\\
17 & \texttt{regular{\textunderscore}plural{\textunderscore}subject{\textunderscore}verb{\textunderscore}agreement{\textunderscore}1} & Subj-V agree\\
18 & \texttt{regular{\textunderscore}plural{\textunderscore}subject{\textunderscore}verb{\textunderscore}agreement{\textunderscore}2} & Subj-V agree \\
19 & \texttt{transitive} & Argument structure\\
20 & \texttt{coordinate{\textunderscore}structure{\textunderscore}constraint{\textunderscore}complex{\textunderscore}left{\textunderscore}branch} & Island effects\\
21 & \texttt{left{\textunderscore}branch{\textunderscore}island{\textunderscore}echo{\textunderscore}question} & Island effects\\
22 & \texttt{wh{\textunderscore}island} & Island effects \\
23 & \texttt{animate{\textunderscore}subject{\textunderscore}passive} & S-selection \\
24 & \texttt{animate{\textunderscore}subject{\textunderscore}trans} & S-selection \\
25 & \texttt{principle{\textunderscore}A{\textunderscore}c{\textunderscore}command} & Binding \\
26 & \texttt{principle{\textunderscore}A{\textunderscore}case{\textunderscore}1} & Binding \\
27 & \texttt{principle{\textunderscore}A{\textunderscore}case{\textunderscore}2} & Binding  \\
28 & \texttt{principle{\textunderscore}A{\textunderscore}domain{\textunderscore}1} & Binding \\
29 & \texttt{principle{\textunderscore}A{\textunderscore}domain{\textunderscore}2} & Binding \\
30 & \texttt{principle{\textunderscore}A{\textunderscore}domain{\textunderscore}3} & Binding\\
31 & \texttt{existential{\textunderscore}there{\textunderscore}object{\textunderscore}raising} & Control/Raising  \\
32 & \texttt{expletive{\textunderscore}it{\textunderscore}object{\textunderscore}raising} & Control/Raising \\
33 & \texttt{only{\textunderscore}npi{\textunderscore}scope} & NPI licensing \\
34 & \texttt{sentential{\textunderscore}negation{\textunderscore}npi{\textunderscore}scope} & NPI licensing\\
35 & \texttt{matrix{\textunderscore}question{\textunderscore}npi{\textunderscore}licensor{\textunderscore}present} & NPI licensing \\
36 & \texttt{npi{\textunderscore}present{\textunderscore}1} & NPI licensing\\
37 & \texttt{npi{\textunderscore}present{\textunderscore}2} & NPI licensing\\
38 & \texttt{only{\textunderscore}npi{\textunderscore}licensor{\textunderscore}present} & NPI licensing \\
39 & \texttt{sentential{\textunderscore}negation{\textunderscore}npi{\textunderscore}licensor{\textunderscore}present} & NPI licensing\\
40 & \texttt{superlative{\textunderscore}quantifiers{\textunderscore}2} & Quantifiers \\
\bottomrule
\end{tabular}
\label{tab:blimp-labels}
\end{table}